\documentclass[runningheads]{llncs}
\usepackage{graphicx}
\usepackage{comment}
\usepackage{amsmath,amssymb} 
\usepackage{color}
\usepackage{booktabs}
\usepackage{makecell}
\usepackage{dirtytalk}
\usepackage{pifont}
\newcommand*\samethanks[1][\value{footnote}]{\footnotemark[#1]}
\usepackage[numbers,sort,compress]{natbib} 
\newcommand{\cmark}{\ding{51}}%
\newcommand{\xmark}{\ding{55}}%

\usepackage[pagebackref=true,breaklinks=true,colorlinks,bookmarks=false]{hyperref}

\usepackage[dvipsnames]{xcolor}

\begin{document}
\pagestyle{headings}
\mainmatter
\def\ECCVSubNumber{1279}  

\newcommand{\samuel}[1]{\textcolor{RubineRed}{[Samuel: #1]}}
\newcommand{\gul}[1]{\textcolor{cyan}{[gul: #1]}}
\newcommand{\lili}[1]{\textcolor{red}{[lili: #1]}}
\newcommand{\datasetName}{BSL-1K}
\newcommand{\posetosign}{Pose$\rightarrow$Sign}
\newcommand{\videotosign}{I3D}

\title{\datasetName: Scaling up co-articulated sign language recognition using mouthing cues}

\titlerunning{\datasetName: Scaling up co-articulated sign language recognition using mouthing cues}
\author{Samuel Albanie$^{1}$\thanks{Equal contribution} \and G\"{u}l Varol$^{1}$\samethanks \and Liliane Momeni$^{1}$ \and Triantafyllos Afouras$^{1}$ \and \\  Joon Son Chung$^{1}$ \and Neil Fox$^{2}$ \and Andrew Zisserman$^{1}$}
\authorrunning{S. Albanie et al.}
\institute{
    $^{1}$Visual Geometry Group, University of Oxford, UK \\
    $^{2}$Deafness, Cognition and Language Research Centre, University College London, UK \\
    \email{
        \{albanie,gul,liliane,afourast,joon,az\}@robots.ox.ac.uk; neil.fox@ucl.ac.uk
    }
}

\maketitle

\begin{abstract}
Recent progress in fine-grained gesture and action classification, and machine translation, point to the possibility of automated sign language 
recognition becoming a reality. A key stumbling block in making progress towards this goal is a lack of
appropriate training data, stemming from the high complexity of sign annotation and a limited supply of qualified annotators.
In this work,
we introduce a new scalable approach to data collection for sign recognition in continuous videos.
We make use of weakly-aligned subtitles for broadcast footage together with a keyword spotting method
 to automatically localise sign-instances for a vocabulary of 1,000 signs in 1,000 hours of video. 
We make the following contributions: (1) We show how to use mouthing cues from signers to obtain high-quality annotations from video data---the result is the \datasetName{} dataset, a collection of British Sign Language (BSL) signs of unprecedented scale; 
(2) We show that we can use \datasetName{} to train strong sign recognition models for co-articulated signs in BSL and that these models additionally form excellent pretraining for other sign languages and benchmarks---we exceed the state of the art on both the MSASL and WLASL benchmarks. Finally, (3)  we propose new large-scale evaluation sets for the tasks of \textit{sign recognition} and \textit{sign spotting} and provide baselines which we hope will serve to stimulate research in this area.

\end{abstract}
\section{Introduction} \label{sec:intro}

With the continual increase in the performance of human action recognition there has been a renewed interest in
the challenge of recognising sign languages such as American Sign Language (ASL),
British Sign Language (BSL), and Chinese Sign Language (CSL).
Although in the past isolated sign recognition has seen some progress, recognition of continuous sign language 
remains extremely challenging~\cite{Camgoz18}. Isolated signs, as in dictionary examples, do not suffer from the \textit{naturally} occurring
complication of co-articulation (i.e. transition motions) between preceding and subsequent signs, making
them visually very different from continuous signing. If we are to recognise ASL
and BSL performed \textit{naturally} by signers, then we need to recognise co-articulated
signs.

Similar problems were faced by Automatic Speech Recognition (ASR) and the solution, as always, was to learn from very
large scale datasets, using a parallel corpus of speech and text. 
In the vision community, a related path was taken with the modern development of automatic lip 
reading: first isolated words were recognised \cite{Chung16}, and later sentences were recognised \cite{Chung17}---in both cases tied to the release of 
large datasets. 
The objective of this paper is to design a scalable \textit{method} to generate large-scale datasets of continuous signing,
for training and testing sign
language recognition, and we demonstrate this for BSL. We start from the perhaps counter-intuitive observation that signers often mouth the word they sign simultaneously, as an additional signal~\cite{Bank2011,sutton-spence_woll_1999,Sutton-Spence2007}, performing similar lip movements as for the spoken word. This differs from mouth gestures which are not derived from the spoken language \cite{crasborn2008frequency}. The mouthing helps disambiguate between different meanings of the same manual sign~\cite{Woll2001} or in some cases simply provides redundancy. In this way, a sign is not only defined by the hand movements and hand shapes, but also by facial expressions and mouth movements~\cite{Cooper11}.

We harness word mouthings to provide a method of automatically
annotating continuous signing. The key idea is to exploit the readily available
and abundant supply of sign-language translated TV broadcasts that consist of an overlaid interpreter performing signs and subtitles that correspond to the audio content. The availability of subtitles means that the annotation task is in essence one of alignment between the words in the subtitle and the mouthings of the overlaid signer. Nevertheless, this is a {\em very} challenging task: a continuous sign may last for only a fraction (e.g.\ 0.5) of a second, whilst the subtitles may last for several seconds and are not synchronised with the signs produced by the signer; the word order of the English need not be the same as the word order of the sign language; the sign may not be mouthed; and furthermore, words may not be signed or may be signed in different ways depending on the context.
For example, the word ``fish'' has a different visual sign depending on referring to the animal or the food, introducing
additional challenges when associating subtitle words to signs.

To detect the mouthings we use
\textit{visual keyword spotting}---the task of determining \textit{whether} and \textit{when} a keyword of interest is uttered by a talking face using \textit{only} visual information---to address the alignment problem described above. Two factors motivate its use: (1) direct lip reading of arbitrary isolated mouthings is a fundamentally difficult task, but searching for a particular known word  within a short temporal window is considerably less challenging; (2) the recent availability of large scale video datasets with aligned audio transcriptions~\cite{Afouras19,Chung16b} now allows for the training of powerful visual keyword spotting models \cite{stafylakis2018zero,sliding-windows,Jha} that, as we show in the experiments, work well for this application.

We make the following contributions: (1) we show how to use visual
keyword spotting to recognise the mouthing cues from signers to obtain
high-quality annotations from video data---the result is the \datasetName{} dataset, a large-scale collection of BSL
(British Sign Language) signs with a 1K sign vocabulary;  (2) We show the
value of \datasetName{} by using it to train strong sign recognition
models for co-articulated signs in BSL and demonstrate that these
models additionally form excellent pretraining for other sign
languages and benchmarks---we exceed the state of the art on both the
MSASL and WLASL benchmarks with this approach; (3) We propose new
evaluation datasets for \textit{sign recognition} and \textit{sign
spotting} and provide baselines for each of these tasks to provide a
foundation for future research\footnote{The project page is at: \url{https://www.robots.ox.ac.uk/~vgg/research/bsl1k/}}.

\section{Related Work} \label{sec:related}

\begin{table}[t]
\caption{\textbf{Summary of previous public sign language datasets:} The \datasetName{} dataset contains, to the best of our
    knowledge, the largest source of annotated sign data in any dataset. It comprises of co-articulated signs
    outside a lab setting.
    }
    \centering
    \resizebox{0.99\linewidth}{!}{
          \begin{tabular}{lcc@{\hskip0.25cm}c@{\hskip0.25cm}c@{\hskip0.25cm}cc}
            \toprule
            Dataset                          & lang & co-articulated & \#signs & \#annos (avg.~per sign) & \#signers  & source    \\
            \midrule
            ASLLVD~\cite{asllvid2008}        & ASL  & \xmark             & 2742    & 9K (3)         & 6                 & lab       \\
            Devisign~\cite{chai2014devisign} & CSL  & \xmark             & 2000    & 24K (12)         & 8                 & lab       \\ 
            MSASL~\cite{Joze19msasl}         & ASL  & \xmark             & 1000    & 25K (25)        & 222               & lexicons, web \\
            WLASL~\cite{Li19wlasl}           & ASL  & \xmark             & 2000    & 21K (11)        & 119                & lexicons, web \\
            \midrule
            S-pot~\cite{viitaniemi14}        & FinSL & \cmark           & 1211    & 4K (3)         & 5                 & lab       \\  
            Purdue RVL-SLLL~\cite{purdue06}  & ASL  & \cmark            & 104     & 2K (19)         & 14             & lab       \\
            Video-based CSL~\cite{Huang2018VideobasedSL} & CSL  & \cmark            & 178      & 25K (140)        & 50               & lab       \\
            SIGNUM~\cite{signum2008}         & DGS  & \cmark            & 455     & 3K  (7)        & 25               & lab       \\
            RWTH-Phoenix~\cite{Koller15cslr,Camgoz18} & DGS  & \cmark            & 1081    & 65K (60)        & 9             & TV        \\
            BSL Corpus~\cite{schembri2013building} & BSL & \cmark &5K &  50K (10) &  249  & lab \\ 
            \midrule
            \textbf{\datasetName{}}                   & BSL  & \cmark            & 1064    & 273K (257)        & 40                & TV \\
            \bottomrule \\
        \end{tabular}
    }
\label{tab:prevdatasets}
\end{table}

\noindent\textbf{Sign language datasets.}
We begin by briefly reviewing public benchmarks for
studying automatic sign language recognition.
Several benchmarks have been proposed for
American~\cite{asllvid2008,Joze19msasl,Li19wlasl,purdue06},
German~\cite{Koller15cslr,signum2008},
Chinese~\cite{chai2014devisign,Huang2018VideobasedSL}, and
Finnish~\cite{viitaniemi14}
sign languages.
BSL datasets, on the other hand,
are scarce. One exception is the ongoing development of the linguistic
corpus~\cite{bslcorpus17,schembri2013building} which provides fine-grained
annotations for the atomic elements of sign production.
Whilst its high annotation quality provides an excellent
resource for sign linguists, the annotations span only
a fraction of the source videos so it is less appropriate
for training current state-of-the-art data-hungry computer vision pipelines.

Tab.~\ref{tab:prevdatasets} presents an overview of publicly
available datasets, grouped according to their provision of \textit{isolated} signs
or \textit{co-articulated} signs.
Earlier datasets have been limited in the size of their video
instances, vocabularies, and signers.
Within the isolated sign datasets,
Purdue RVL-SLLL~\cite{purdue06} has a limited vocabulary of 104 signs
(ASL comprises more than 3K signs in total~\cite{valli2005gallaudet}).
ASLLVD~\cite{asllvid2008} has only 6 signers.
Recently, MSASL~\cite{Joze19msasl} and WLASL~\cite{Li19wlasl}
large-vocabulary isolated sign datasets
have been released
with 1K and 2K signs, respectively.
The videos are collected from lexicon databases
and other instructional videos on the web.

Due to the difficulty of annotating co-articulated signs
in long videos,
continuous datasets have been limited in their vocabulary,
and most of them have been recorded in lab
settings~\cite{Huang2018VideobasedSL,signum2008,purdue06}.
RWTH-Phoenix~\cite{Koller15cslr} is one of the few realistic
datasets that supports training complex models based on
deep neural networks.
A recent extension also allows studying sign language translation~\cite{Camgoz18}.
However, the
videos in~\cite{Camgoz18,Koller15cslr} are only from weather broadcasts,
restricting the domain of discourse.
In summary, the main constraints of the previous datasets are one or more of the following: (i) they are limited in size, (ii) they have a large total vocabulary but only of isolated signs, or (iii) they consist of natural co-articulated signs but cover a limited domain of discourse.
The \datasetName{} dataset provides a considerably greater number of
annotations than all previous public sign language datasets,
and it does so in the co-articulated setting for a large domain of discourse.

\noindent\textbf{Sign language recognition.}
Early work on sign language recognition
focused on hand-crafted features
computed for hand shape and motion~\cite{Farhadi07,Fillbrandt2003,Starner95,Tamura88}.
Upper body and hand pose have then
been widely used as part of the recognition
pipelines~\cite{Buehler09,Camgoz17,Cooper2011,Ong2012,Pfister14}.
Non-manual features such as
face~\cite{Farhadi07,Koller15cslr,Nguyen2008},
and mouth~\cite{Antonakos2015,koller2014,koller15:mouth}
shapes are relatively less considered.
For sequence modelling of signs, 
HMMs~\cite{Farhadi2006,Ulrich2008,Forster2013,Starner95}, and
more recently LSTMs~\cite{Camgoz17,Huang2018VideobasedSL,Ye18,zhou2020spatialtemporal},
have been utilised.
Koller~et~al.~\cite{Koller17ReSign} present a hybrid approach based on CNN-RNN-HMM
to iteratively re-align sign language videos to the sequence of sign annotations.
More recently 3D CNNs have been adopted due to
their representation capacity for  spatio-temporal
data~\cite{Bilge19ZS,camgoz2016:3dconv,Huang15,Joze19msasl,Li19wlasl}.
Two recent concurrent works~\cite{Joze19msasl,Li19wlasl} showed that I3D models~\cite{Carreira2017}
significantly outperform
their  pose-based counterparts.
In this paper, we confirm the success of I3D models,
while also showing improvements using pose distillation as pretraining.
There have been efforts to use sequence-to-sequence translation models for 
sign language translation~\cite{Camgoz18}, though this has been limited to the weather discourse of RWTH-Phoenix,
and the method is limited by the size of the training set. The recent work of~\cite{li2020transferring} localises signs in continuous news footage to improve an isolated sign classifier.

In this work, we utilise mouthings to localise signs in weakly-supervised videos.
Previous work~\cite{Buehler09,Cooper2009,Pfister14,Chung16b} has used weakly aligned subtitles
as a source of training data, and both one-shot~\cite{Pfister14}
(from a visual dictionary) and zero-shot~\cite{Bilge19ZS} (from  a textual description) have also
been used. Though no previous work, to our knowledge, has put these ideas together.
The sign spotting problem was formulated in~\cite{ong2014,viitaniemi14}.

\noindent \textbf{Using the mouth patterns.}
The mouth has several roles in sign language that can be grouped into
spoken components (mouthings) and oral components (mouth gestures)~\cite{Woll2001}.
Several works focus on recognising mouth shapes~\cite{Antonakos2015,koller15:mouth}
to recover mouth gestures.
Few works~\cite{koller2014,koller2014:mouthings} attempt to 
recognise mouthings in sign language data by focusing on a few categories of
visemes, i.e., visual correspondences
of phonemes in the lip region~\cite{Fisher1968}.
Most closely related to our 
work,~\cite{pfister2013large} similarly searches subtitles of broadcast footage 
and uses the mouth as a cue to improve alignment between the subtitles and the 
signing. Two key differences between our work and theirs are: (1) we achieve 
precise localisation through keyword spotting, whereas they only use an open/closed 
mouth classifier to reduce the number of candidates for a given sign; (2) 
scale---we gather signs over 1,000 hours of signing (in contrast to the 30 
hours considered in~\cite{pfister2013large}).

\section{Learning Sign Recognition with Automatic Labels} \label{sec:method}

\begin{figure}[t]
    \centering
    \includegraphics[width=.9\textwidth,trim={0cm 0cm 0 0cm},clip]{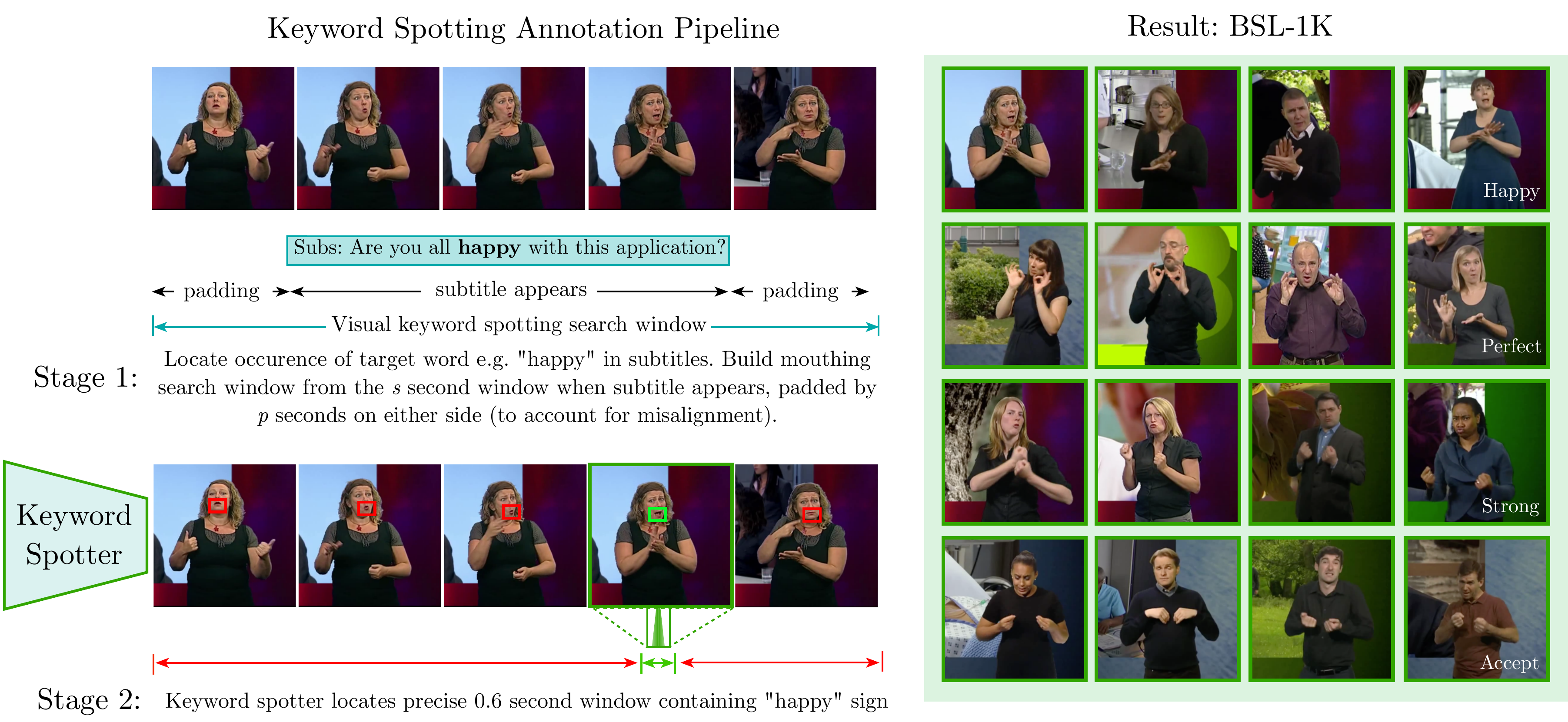}
    \caption{\textbf{Keyword-driven sign annotation:} (Left, the annotation pipeline): Stage~1: for a given target sign (e.g. ``happy'') each occurrence of the word in the subtitles provides a candidate temporal window when the sign may occur (this is further padded by several seconds on either side to account for misalignment of subtitles and signs); Stage~2: a keyword spotter uses the mouthing of the signer to perform precise localisation of the sign within this window. (Right): Examples from the \datasetName{} dataset---produced by applying keyword spotting for a vocabulary of 1K words.}
    \label{fig:pipeline}
\end{figure}

In this section, we describe the process used to collect \datasetName{}, a large-scale dataset of BSL signs.  An overview of the approach is provided in Fig.~\ref{fig:pipeline}. In Sec.~\ref{subsec:coarse-loc}, we describe how large numbers of video clips that are likely to contain a given sign are sourced from public broadcast footage using subtitles; in Sec.~\ref{subsec:keyword-loc}, we show how automatic keyword spotting can be used to precisely localise specific signs to within a fraction of a second; in Sec.~\ref{subsec:building-bs1-1k}, we apply this technique to efficiently annotate a large-scale dataset with a vocabulary of 1K signs. 

\subsection{Finding probable signing windows in public broadcast footage \label{subsec:coarse-loc}}

The source material for the dataset comprises 1,412 episodes of publicly broadcast TV programs produced by the BBC which contains 1,060 hours of continuous BSL signing.  The episodes cover a wide range of topics: medical dramas, history and nature documentaries, 
cooking shows and programs covering gardening, business and travel. The signing represents a translation (rather than a transcription) of the content and is produced by a total of forty professional BSL interpreters. The signer occupies a fixed region of the screen and is cropped directly from the footage. A full list of the TV shows that form \datasetName{} can be found in
Appendix~\ref{app:subsec:episodes}.
In addition to videos, these episodes are accompanied by subtitles (numbering approximately 9.5 million words in total).  To locate temporal windows in which instances of signs are likely to occur within the source footage, we first identify a candidate list of words that: (i) are present in the subtitles; (ii) have entries in both BSL 
signbank\footnote{\url{https://bslsignbank.ucl.ac.uk/}} and sign BSL\footnote{\url{https://www.signbsl.com/}}, two online dictionaries of isolated signs (to ensure that we query words that have valid mappings to signs).  The result is an initial vocabulary of 1,350 words, which are used as queries for the keyword spotting model to perform sign localisation---this process is described next.

\subsection{Precise sign localisation through visual keyword spotting \label{subsec:keyword-loc}}

By searching the content of the subtitle tracks for instances of words in the initial vocabulary, we obtain a set of candidate temporal windows in which instances of signs may occur.  However, two factors render these temporal proposals extremely noisy: (1)  the presence of a word in the subtitles does not guarantee its presence in the signing; (2) even for subtitled words that are signed, we find through inspection that their appearance in the subtitles can be misaligned with the sign itself by several seconds.

To address this challenge, we turn to \textit{visual keyword spotting}.  Our goal is to detect and precisely localise the presence of a sign by identifying its \say{spoken components}~\cite{sutton-spence_woll_1999} within a temporal sequence of mouthing patterns.  Two hypotheses underpin this approach: (a) that mouthing provides a strong  localisation signal for signs as they are produced; (b) that this mouthing occurs with sufficient frequency to form a useful localisation cue.  Our method is motivated by studies in the Sign Linguistics literature which find that spoken components frequently serve to identify signs---this occurs most prominently when the mouth pattern is used to distinguish between manual homonyms\footnote{These are signs that use identical hand movements (e.g. \say{king} and \say{queen}) whose meanings are distinguished by mouthings.} (see~\cite{sutton-spence_woll_1999} for a detailed discussion).   However, even if these hypotheses hold, the task remains extremely challenging---signers typically do not mouth continuously and the mouthings that are produced may only correspond to a portion of the word~\cite{sutton-spence_woll_1999}.  For this reason, existing lip reading approaches cannot be used directly (indeed, an initial exploratory experiment we conducted with the state-of-the-art lip reading model of~\cite{Afouras19} achieved zero recall on five-hundred randomly sampled sentences of signer mouthings from the BBC source footage). 

The key to the effectiveness of visual keyword spotting is that rather than solving the general problem of lip reading, it solves the much easier problem of identifying a single token from a small collection of candidates within a short temporal window.  In this work, we use the subtitles to construct such windows.  The pipeline for automatic sign annotations therefore consists of two stages (Fig.~\ref{fig:pipeline}, left): (1) For a given target sign e.g. ``happy'', determine the times of all occurrences of this sign in the subtitles accompanying the video footage. The subtitle time provides a short window during which the word was spoken, but not necessarily when its corresponding sign is produced in the translation.  We therefore extend this candidate window by several seconds to increase the likelihood that the sign is present in the sequence.  We include ablations to assess the influence of this padding process in Sec.~\ref{sec:experiments} and determine empirically that padding by four seconds on each side of the subtitle represents a good choice. (2)~The resulting temporal window is then provided, together with the target word, to a keyword spotting model (described in detail in Sec.~\ref{subsec:visual-kws}) which estimates the probability that the sign was mouthed at each time step (we apply the keyword spotter with a stride of $0.04$ seconds---this choice is motivated by the fact that the source footage has a frame rate of 25fps).  When the keyword spotter asserts with high confidence that it has located a sign, we take the location of the peak posterior probability as an anchoring point for one endpoint of a 0.6~second window (this value was determined by visual inspection to be sufficient for capturing individual signs).  The peak probability is then converted into a decision about whether a sign is present using a threshold parameter. To build the \datasetName{} dataset, we select a value of 0.5 for this parameter after conducting experiments (reported in  Tab.~\ref{table:mouthing}) to assess its influence on the downstream task of sign recognition performance.

\begin{figure}[t]
    \centering
    \includegraphics[width=0.9\textwidth,trim={4.5cm 0cm 4.5cm 0cm},clip]{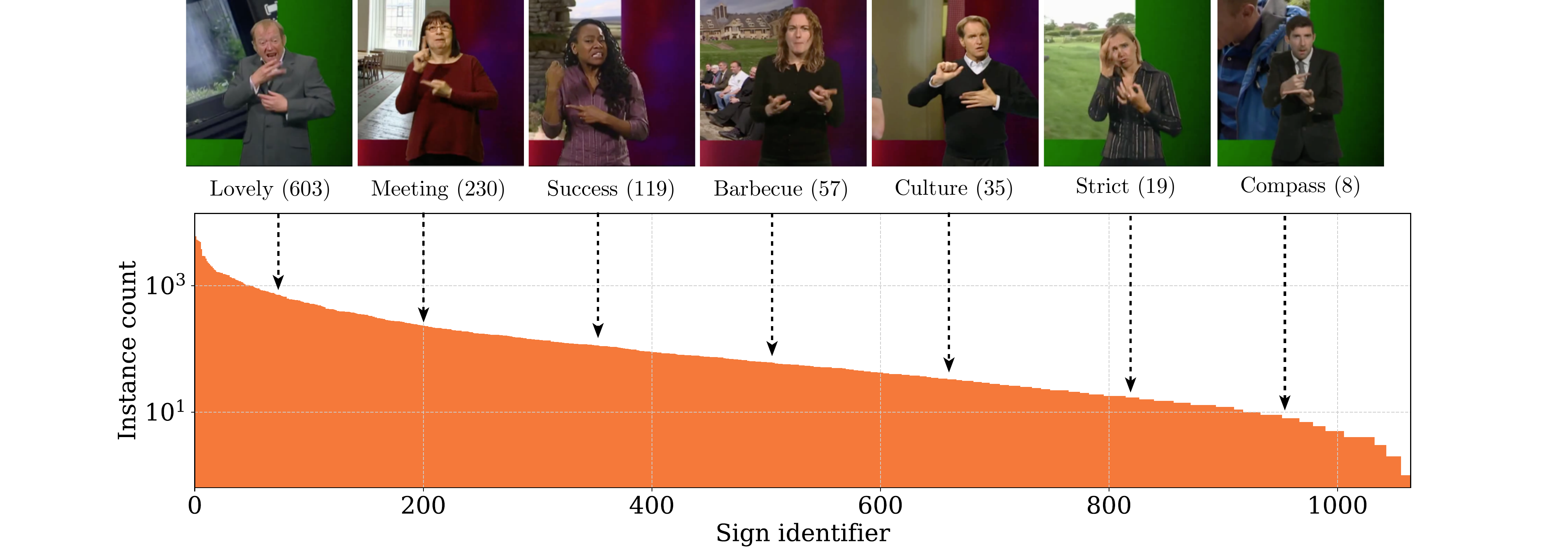}
    \caption{\textbf{\datasetName{} sign frequencies:} Log-histogram of instance counts for the 1,064 words constituting the \datasetName{} vocabulary, together with example signs. The long-tail distribution reflects the \textit{real} setting in which some signs are more frequent than others.}
    \label{fig:word-distribution}
\end{figure}

\subsection{\datasetName{} dataset construction and validation \label{subsec:building-bs1-1k}}

Following the sign localisation process described above, we obtain approximately 280k localised signs from a set of 2.4 million candidate subtitles.  To ensure that the dataset supports study of signer-independent sign recognition, we then compute face embeddings (using an SENet-50~\cite{hu2019squeeze} architecture trained for verification on the VGGFace2 dataset~\cite{Cao18}) to group the episodes according to which of the forty signers they were translated by.  We partition the data into three splits, assigning thirty-two signers for training, four signers for validation and four signers for testing.  We further sought to include an equal number of hearing and non-hearing signers (the validation and test sets both contain an equal number of each, the training set is approximately balanced with 13 hearing, 17 non-hearing and 2 signers whose deafness is unknown). 
We then perform a further filtering step on the vocabulary to ensure that each word included in the dataset is represented with high confidence (at least one instance with confidence 0.8) in the training partition, which produces a final dataset vocabulary of 1,064 words (see Fig.~\ref{fig:word-distribution} for the distribution and
Appendix~\ref{app:subsec:vocab}
for the full word list).

\begin{table}[t]
    \setlength{\tabcolsep}{8pt}
    \centering
\caption{\textbf{Statistics of the proposed \datasetName{} dataset:} The \textit{Test-(manually verified)} split represents a sample from the Test-(automatic) split annotations that have been verified by human annotators (see Sec.~\ref{subsec:building-bs1-1k} for details).
}
    \resizebox{0.8\linewidth}{!}{
        \begin{tabular}{lccc}
            \toprule
            Set & sign vocabulary & sign annotations & number of signers \\
            \midrule
            Train & 1,064 & 173K & 32 \\
            Val & 1,049 & 36K & 4 \\
            \midrule
            Test-(automatic) & 1,059 & 63K & 4 \\
            Test-(manually verified) & 334 & 2103 & 4 \\
            \bottomrule \\
        \end{tabular}
    }
\label{tab:statistics}
\end{table}

\noindent \textbf{Validating the automatic annotation pipeline.} One of the key hypotheses underpinning this work is that keyword spotting is capable of correctly locating signs. We first verify this hypothesis by presenting a randomly sampled subset of the test partition to a native BSL signer, who was asked to assess whether the short temporal windows produced by the keyword spotting model with high confidence (each 0.6 seconds in duration) contained correct instances of the target sign.  A screenshot of the annotation tool developed for this task is provided in
Fig.~\ref{app:fig:verification-tool}.
A total of 1k signs were included in this initial assessment, of which 70\% were marked as correct, 28\% were marked as incorrect and 2\% were marked as uncertain, validating the key idea behind the annotation pipeline.  
Possible reasons for incorrect marks include:
BSL mouthing patterns are not always identical to 
spoken English and mouthings many times do not represent the full word
(e.g., ``fsh'' for ``finish'')~\cite{sutton-spence_woll_1999}.

\noindent \textbf{Constructing a manually verified test set.} To construct a high quality, human verified test set and to maximise yield from the annotators, we started from a collection of sign predictions where the keyword model was highly confident (assigning a peak probability of greater than 0.9) yielding 5,826 sign predictions.   Then, in addition to the validated 980 signs (corrections were provided as labels for the signs marked as incorrect and uncertain signs were removed), we further expanded the verified test set with non-native (BSL level 2 or above) signers who annotated a further 2k signs.  We found that signers with lower levels of fluency were able to confidently assert that a sign was correct for a portion of the signs (at a rate of around 60\%), but also annotated a large number of signs as ``unsure'', making it challenging to use these annotations as part of the validation test for the effectiveness of the pipeline.  Only signs marked as correct were included into the final verified test set, which ultimately comprised 2,103 annotations covering 334 signs from the 1,064 sign vocabulary.  The statistics of each partition of the dataset are provided in Tab.~\ref{tab:statistics}.   All experimental test set results in this paper refer to performance on the verified test set (but we retain the full automatic test set, which we found to be useful for development).

In addition to the keyword spotting approach described above, we explore techniques for further dataset expansion based on other cues in
Appendix~\ref{app:subsec:othercues}.

\section{Models and Implementation Details}

In this section, we first describe the visual keyword spotting model used to collect signs from mouthings (Sec.~\ref{subsec:visual-kws}). Next, we provide details of
the model architecture for sign recognition and spotting (Sec.~\ref{subsec:method:training}). Lastly, we describe a method for obtaining a good initialisation for the sign recognition model (Sec.~\ref{subsec:method:posedistillation}).

\subsection{Visual keyword spotting model \label{subsec:visual-kws}}

We use the improved visual-only keyword spotting model of Stafylakis et al.~\cite{stafylakis2018zero} from \cite{Momeni2020kws} (referred to in their paper as ``P2G \cite{stafylakis2018zero} baseline"), provided by the authors. The model of  \cite{stafylakis2018zero} combines visual features with a fixed-length keyword embedding to determine whether a user-defined keyword is present in an input video clip. The performance of \cite{stafylakis2018zero} is improved in \cite{Momeni2020kws} by switching the keyword encoder-decoder from grapheme-to-phoneme (G2P) to phoneme-to-grapheme (P2G). 

In more detail, the model consists of four stages: (i) visual features are first extracted from the sequence of face-cropped image frames from a clip (this is performed using a $512 \times 512$ SSD architecture~\cite{liu2016ssd} trained for face detection on WIDER faces~\cite{yang2016wider}), (ii) a fixed-length keyword representation is built using a P2G encoder-decoder, (iii) the visual and keyword embeddings are concatenated and passed through BiLSTMs, (iv) finally, a sigmoid activation is applied on the output to approximate the posterior probability that the keyword occurs in the video  clip  for  each  input  frame.  If  the  maximum  posterior  over  all  frames  is greater than a threshold, the clip is predicted to contain the keyword. The predicted location of the keyword is the position of the maximum posterior. Finally, non-maximum suppression is run with a temporal window of 0.6 seconds over the untrimmed source videos to remove duplicates.

\subsection{Sign recognition model} \label{subsec:method:training}
We employ a spatio-temporal convolutional neural network
architecture that takes a multiple-frame video as input,
and outputs class probabilities over sign categories.
Specifically, we follow the I3D architecture~\cite{Carreira2017}
due to its success on action recognition benchmarks, as well
as its recently observed success on sign recognition
datasets~\cite{Joze19msasl,Li19wlasl}.
To retain computational efficiency, we only use an RGB stream.
The model is trained on 16-frame consecutive frames (i.e., 0.64 sec for 25fps),
as \cite{Buehler09,pfister2013large,viitaniemi14}
observed that co-articulated signs last roughly for 13 frames.
We resize our videos to have a spatial resolution of $224 \times 224$.
For training, we randomly subsample a fixed-size, temporally contiguous input from the spatio-temporal volume
to have $16 \times 224 \times 224$ resolution
in terms of number of frames, width, and height, respectively.
We minimise the cross-entropy loss using SGD with momentum (0.9)
with mini-batches of size 4, and an initial learning rate of $10^{-2}$
with a fixed schedule. The learning rate is decreased twice with a factor of $10^{-1}$
at epochs 20 and 40. We train for 50 epochs. Colour, scale, and horizontal flip
augmentations are applied on the input video.
When pretraining is used (e.g. on Kinetics-400~\cite{Carreira2017} or on other data where specified), we replace the last linear layer with the dimensionality of our classes, and fine-tune all network parameters (we observed that freezing part of the model is suboptimal). Finally, we apply dropout on the classification layer with a probability of 0.5.

At test time, we perform centre-cropping and apply a sliding window
with a stride of 8 frames before averaging the classification scores to obtain a video-level
prediction.

\subsection{Video pose distillation} \label{subsec:method:posedistillation}
Given the significant focus on pose estimation in the sign
language recognition literature, we investigate how explicit pose modelling
can be used to improve the I3D model.  To this end, we define a \textit{pose distillation} network that takes in a sequence of 16 consecutive frames, but rather than predicting sign categories, the 1024-dimensional (following average pooling) embedding produced by the network is used to regress the poses of individuals appearing in each of the frames of its input.  In more detail, we assume a single individual per-frame (as is the case in cropped sign translation footage) and task the network with predicting 130 human pose keypoints (18 body, 21 per hand, and 70 facial) produced by an OpenPose~\cite{cao2018openpose} model (trained on COCO~\cite{lin2014microsoft}) that is evaluated per-frame. The key idea is that, in order to effectively predict pose across multiple frames from a single video embedding, the model is encouraged to encode information not only about pose, but also descriptions of relevant dynamic gestures.  The model is trained on a portion of the \datasetName{} training set (due to space constraints, further details of the model architecture and training procedure are provided in
Appendix~\ref{app:sec:posedistillation}).

\section{Experiments} \label{sec:experiments}
We first provide several ablations on our sign recognition
model to answer questions such as which cues are important,
and how to best use human pose.
Then, we present baseline results for sign recognition and sign spotting,
with our best model.
Finally, we compare to the state of the art on ASL benchmarks
to illustrate the benefits of pretraining on our data.

\subsection{Ablations for the sign recognition model}
In this section, we evaluate our sign language recognition approach
and investigate (i) the effect of mouthing score threshold, (ii) the comparison to pose-based approaches,
(iii) the contribution of multi-modal cues, and
(iv) the video pose distillation.
Additional ablations about the influence of
the search window size for the keyword spotting (Appendix~\ref{app:subsec:kwswindow})
and
the temporal extent of the automatic annotations (Appendix~\ref{app:subsec:numframes})
can be found in
the appendix.

\noindent\textbf{Evaluation metrics.}
Following~\cite{Joze19msasl,Li19wlasl}, we report
both top-1 and top-5 classification accuracy,
mainly due to ambiguities in signs which can be resolved
in context. 
Furthermore, we adopt both per-instance and per-class
accuracy metrics. Per-instance accuracy is computed over all
test instances. Per-class accuracy refers to the average over the
sign categories present in the test set.
We use this metric due to the unbalanced nature of the datasets.

\noindent\textbf{The effect of the mouthing score threshold.}
The keyword spotting method, being a binary classification model,
provides a confidence score, which we threshold to obtain
our 
automatically annotated video clips. Reducing this threshold yields an increased number of sign instances at the cost of a potentially noisier set of annotations.
We denote the training set defined by a mouthing threshold 0.8
as \datasetName{}$_{m.8}$.
In Tab.~\ref{table:mouthing},
we show the effect of changing this hyper-parameter
between a low- and high-confidence model with 0.5 and 0.8 mouthing
thresholds, respectively.
The larger set of training samples obtained with a threshold of 0.5 provide the best performance.
For the remaining ablations, we use the smaller \datasetName{}$_{m.8}$ training
set for faster iterations, and return to the larger \datasetName{}$_{m.5}$ set for training the final model.

\begin{table}[t]
    \setlength{\tabcolsep}{8pt}
    \centering
    \caption{
    \textbf{Trade-off between training noise vs. size:} Training (with Kinetics initialisation)
    on the full training set \datasetName{}$_{m.5}$ versus
    the subset \datasetName{}$_{m.8}$,
    which correspond to a mouthing score threshold of 0.5 and 0.8,
    respectively.
    Even when noisy, with the 0.5 threshold,
    mouthings provide automatic annotations that allow
    supervised training at scale, resulting in 70.61\%
    accuracy on the manually validated test set. 
    }
    \resizebox{0.9\linewidth}{!}{
        \begin{tabular}{lr|cccc}
            \toprule
            & & \multicolumn{2}{c}{per-instance} & \multicolumn{2}{c}{per-class} \\
            Training data & \#videos & top-1 & top-5 & top-1 & top-5  \\
            \midrule
            \datasetName$_{m.8}$ (mouthing$\geq$0.8) & 39K  & 69.00 & 83.79 & 45.86 & 64.42 \\ 
            \datasetName$_{m.5}$ (mouthing$\geq$0.5) & 173K & \textbf{70.61} & \textbf{85.26} & \textbf{47.47} & \textbf{68.13} \\
            \bottomrule
        \end{tabular}
    }
    \label{table:mouthing}
\end{table}

\noindent\textbf{Pose-based model versus I3D.} 
We next verify that I3D is a suitable model for sign language recognition
by comparing it to a pose-based approach.
We implement \posetosign{}, which
follows a 2D ResNet architecture~\cite{he2016deep} that
operates on $3\times 16 \times P$ dimensional dynamic pose images,
where $P$ is the number of keypoints. In our experiments,
we use OpenPose~\cite{cao2018openpose} (pretrained on COCO~\cite{lin2014microsoft}) to extract
18 body, 21 per hand, and 70 facial keypoints. We use
16-frame inputs to make it comparable to the I3D counterpart.
We concatenate the estimated normalised
$xy$ coordinates of a keypoint with its
confidence score to obtain the 3 channels.
In Tab.~\ref{table:pose2sign}, we see that
I3D significantly outperforms the explicit 2D pose-based
method (65.57\% vs 49.66\% per-instance accuracy).
This conclusion is in accordance with the recent findings
of~\cite{Joze19msasl,Li19wlasl}.

\noindent\textbf{Contribution of individual cues.}
We carry out two set of experiments to determine
how much our sign recognition model relies on signals from the
mouth and face region
versus the manual features from the body and hands:
(i) using \posetosign{}, which takes as input
the 2D keypoint locations over several frames,
(ii) using \videotosign{}, which takes as input
raw video frames. 
For the pose-based model, we train with only 70 facial keypoints,
60 body\&hand keypoints, or with the combination.
For I3D, we use the pose estimations to mask the pixels outside
of the face bounding box, to mask the mouth region, or
use all the pixels from the videos.
The results are summarised in Tab.~\ref{table:pose2sign}.
We observe that using only the face provides a strong baseline,
suggesting that mouthing is a strong cue for recognising signs,
e.g., 42.23\% for I3D.
However, using all the cues, including body and hands (65.57\%),
significantly outperforms using individual modalities.

\begin{table}[t]
    \setlength{\tabcolsep}{8pt}
    \centering
    \caption{\textbf{Contribution of individual cues:} We compare I3D (pretrained on Kinetics) with a keypoint-based baseline
        both trained and evaluated on a subset of
        \datasetName{}$_{m.8}$, where we have the pose estimates. We also quantify
        the contribution of the body\&hands and the face regions. We see that
        significant information can be attributed to both types of cues, and
        the combination performs the best.}
    \resizebox{0.79\linewidth}{!}{
        \begin{tabular}{lllcccc}
            \toprule
                 & & & \multicolumn{2}{c}{per-instance} & \multicolumn{2}{c}{per-class} \\
                           & body\&hands & face & top-1 & top-5 & top-1 & top-5 \\
            \midrule
            \posetosign{} (70 points)  & \xmark & \cmark & 24.41 & 47.59 &  9.74 & 25.99 \\
            \posetosign{} (60 points)  & \cmark & \xmark & 40.47 & 59.45 & 20.24 & 39.27 \\
            \posetosign{} (130 points) & \cmark & \cmark & \textbf{49.66} & \textbf{68.02} & \textbf{29.91} & \textbf{49.21} \\
            \midrule
            \videotosign{} (face-crop)    & \xmark & \cmark & 42.23 & 69.70 & 21.66 & 50.51 \\
            \videotosign{} (mouth-masked) & \cmark & \xmark & 46.75 & 66.34 & 25.85 & 48.02 \\
            \videotosign{} (full-frame)   & \cmark & \cmark & \textbf{65.57} & \textbf{81.33} & \textbf{44.90} & \textbf{64.91} \\
            \bottomrule
        \end{tabular}
    }
    \label{table:pose2sign}
\end{table}
\begin{table}[t]
    \setlength{\tabcolsep}{8pt}
    \centering
    \caption{\textbf{Effect of pretraining} the I3D model on various tasks before
        fine-tuning for sign recognition on \datasetName{}$_{m.8}$.
        Our dynamic pose features learned on 16-frame videos
        provide body-motion-aware cues and outperform other pretraining
        strategies.
        }
    \resizebox{0.79\linewidth}{!}{
        \begin{tabular}{ll|cccc}
            \toprule
            \multicolumn{2}{c|}{Pretraining} & \multicolumn{2}{c}{per-instance} & \multicolumn{2}{c}{per-class} \\
            Task & Data & top-1 & top-5 & top-1 & top-5  \\
            \midrule
            Random init. & - & 39.80 & 61.01 & 15.76 & 29.87 \\
            Gesture recognition & Jester~\cite{Materzynska_2019_ICCV} & 46.93 & 65.95 & 19.59 & 36.44 \\ 
            Sign recognition & WLASL~\cite{Li19wlasl} & 69.90 & 83.45 & 44.97 & 62.73 \\ 
            Action recognition & Kinetics~\cite{Carreira2017} & 69.00 & 83.79 & 45.86 & 64.42 \\ 
            Video pose distillation & Signers & \textbf{70.38} & \textbf{84.50} & \textbf{46.24} & \textbf{65.31} \\ 
            \bottomrule
        \end{tabular}
    }
    \label{table:pretraining}
\end{table}

\noindent\textbf{Pretraining for sign recognition.} Next we investigate different forms of pretraining for the I3D model.  In Tab.~\ref{table:pretraining}, we compare the performance of a model trained with random initialisation (39.80\%), fine-tuning from gesture recognition (46.93\%), sign recognition (69.90\%), and action recognition (69.00\%). Video pose distillation provides a small boost over the other pretraining strategies (70.38\%), suggesting that it is an effective way to force the I3D model to pay attention to the dynamics of the human keypoints, which is relevant for sign recognition. 

\subsection{Benchmarking sign recognition and sign spotting}

\begin{figure}[t]
    \centering
    \includegraphics[width=.9\textwidth,trim={0cm 0cm 0 0cm},clip]{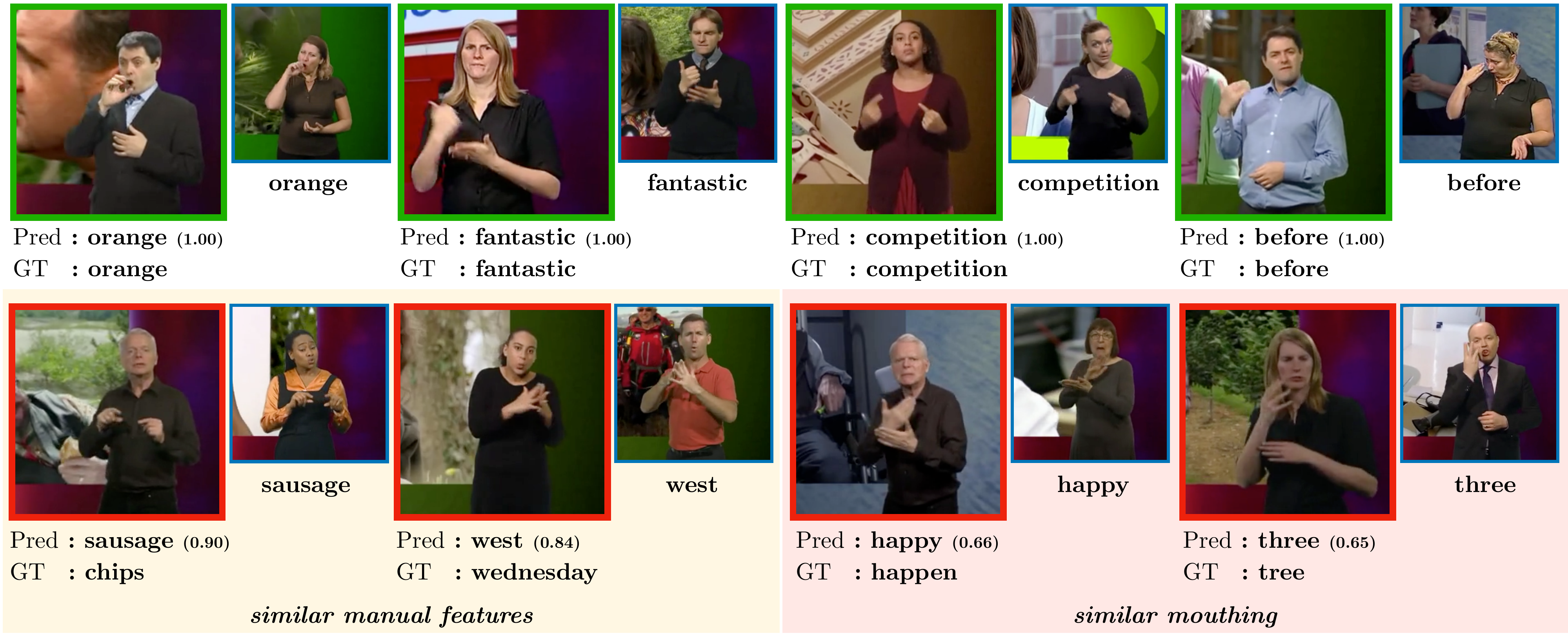}
    \caption{\textbf{Qualitative analysis:} We present results of our sign recognition model on \datasetName{}
    for success (top) and failure (bottom) cases, together with their confidence scores
    in parentheses.
    To the right of each example, we show
    a random training sample for the predicted sign (in small).
    We observe that failure modes are commonly due to high visual similarity in the gesture (bottom-left)
    and mouthing (bottom-right).
    }
    \label{fig:qualitative}
\end{figure}

\begin{table}[t]
    \setlength{\tabcolsep}{8pt}
    \centering
    \caption{\textbf{Benchmarking:} We benchmark our best sign recognition model
    (trained on \datasetName{}$_{m.5}$, initialised with pose distillation, with 4-frame
    temporal offsets) 
    for sign recognition and sign spotting task to establish strong baselines on \datasetName{}.
    }
    \resizebox{0.5\linewidth}{!}{
        \begin{tabular}{lcccc}
            \toprule
             & \multicolumn{2}{c}{per-instance} & \multicolumn{2}{c}{per-class} \\
             & top-1 & top-5 & top-1 & top-5  \\
            \midrule
            Sign Recognition  & 75.51 & 88.83 & 52.76 & 72.14 \\
            \bottomrule
        \end{tabular}
    } 
    \resizebox{0.34\linewidth}{!}{
        \begin{tabular}{lc}
            \toprule
            & \multicolumn{1}{c}{mAP} \\
            & \multicolumn{1}{c}{(334 sign classes)} \\
            \midrule
            Sign Spotting & 0.159 \\
            \bottomrule
        \end{tabular}
    }
    \label{table:best}
\end{table}

Next, we combine the parameter choices suggested by each of our ablations to establish baseline performances on the \datasetName{}
dataset for two tasks: (i) sign recognition, (ii) sign spotting. Specifically,
the model
comprises an I3D architecture trained on \datasetName{}$_{m.5}$ with pose-distillation as initialisation and random temporal offsets of up to 4 frames around the sign during training (the ablation studies for this temporal augmentation parameter are included in
Appendix~\ref{app:subsec:numframes}).
The sign recognition evaluation protocol follows the experiments conducted in the ablations, the sign spotting protocol is described next.

\noindent\textbf{Sign spotting.}
Differently from sign recognition, in which the objective is to classify a pre-defined temporal segment into a category from a given vocabulary, \textit{sign spotting} aims to locate all instances of a particular sign within long sequences of untrimmed footage, enabling applications such as content-based search and efficient indexing of signing videos for which subtitles are not available.  The evaluation protocol for assessing sign spotting on \datasetName{} is defined as follows: for each sign category present amongst the human-verified test set annotations (334 in total), windows of 0.6-second centred on each verified instance are marked as positive and all other times within the subset of episodes that contain at least one instance of the sign are marked as negative.  To avoid false penalties at signs that were not discovered by the automatic annotation process, we exclude windows of 8 seconds of footage centred at each location in the original footage at which the target keyword appears in the subtitles, but was not detected by the visual keyword spotting pipeline. In aggregate this corresponds to locating approximately one positive instance of a sign in every 1.5 hours of continuous signing negatives.  A sign is considered to have been correctly spotted if its temporal overlap with the model prediction exceeds an IoU (intersection-over-union) of 0.5, and we report mean Average Precision (mAP) over the 334 sign classes as the metric for performance.

We report the performance of our strongest model for both the sign recognition and sign spotting benchmarks in Tab.~\ref{table:best}. In Fig.~\ref{fig:qualitative}, we provide some qualitative results from our sign recognition method and observe some modes of failure which are driven by strong visual similarity in sign production.

\subsection{Comparison with the state of the art on ASL benchmarks} \label{subsec:exp:sota}
\datasetName{}, being significantly larger than the recent WLASL~\cite{Li19wlasl} and MSASL~\cite{Joze19msasl} 
benchmarks, can be used for pretraining I3D models to provide strong
initialisation for other datasets. Here, we transfer the features from BSL to ASL,
which are two distinct sign languages.

As models from \cite{Joze19msasl} were not publicly available,
we first reproduce the I3D Kinetics pretraining baseline with our implementation
to achieve fair comparisons. We use 64-frame inputs as isolated signs in these datasets are significantly
slower than co-articulated signs. We then train I3D from \datasetName{} pretrained features.
Tab.~\ref{table:aslsota} compares pretraining on Kinetics versus
our \datasetName{} data. \datasetName{} provides a significant boost
in the performance, outperforming the state-of-the-art results
(46.82\% and 64.71\% top-1 accuracy). Find additional details, as well
as similar experiments
on co-articulated datasets in
Appendix~\ref{app:subsec:transfer}.

\begin{table}[t]
    \setlength{\tabcolsep}{8pt}
    \centering
    \caption{\textbf{Transfer to ASL:} Performance on American Sign Language (ASL) datasets with and without pretraining on our data. I3D
    results are reported from the original papers for MSASL~\cite{Joze19msasl}
    and WLASL~\cite{Li19wlasl}. I3D$\dagger$ denotes
    our implementation and training, adopting the hyper-parameters from \cite{Joze19msasl}.
    We show that our features provide good initialisation, even if it is trained
    on BSL.
    }
    \resizebox{0.99\linewidth}{!}{
        \begin{tabular}{llcccccccc}
            \toprule
            & & \multicolumn{4}{c}{WLASL~\cite{Li19wlasl}} & \multicolumn{4}{c}{MSASL~\cite{Joze19msasl}} \\
            & & \multicolumn{2}{c}{per-instance} & \multicolumn{2}{c}{per-class} &
            \multicolumn{2}{c}{per-instance} & \multicolumn{2}{c}{per-class} \\
            & pretraining & top-1 & top-5 & top-1 & top-5 & top-1 & top-5 & top-1 & top-5 \\
            \midrule
            I3D \cite{Joze19msasl} & Kinetics & - & - & - & -  & - & - & 57.69 & 81.08 \\
            I3D \cite{Li19wlasl} & Kinetics & 32.48 & 57.31 & - & -  & - & - & - & - \\
            I3D$\dagger$ & Kinetics & 40.85 & 74.10 & 39.06 & 73.33 & 60.45 & 82.05 & 57.17 & 80.02 \\
            I3D & \datasetName{} & \textbf{46.82} & \textbf{79.36} & \textbf{44.72} & \textbf{78.47} & \textbf{64.71} & \textbf{85.59} & \textbf{61.55} & \textbf{84.43} \\
            \bottomrule
        \end{tabular}
    }
    \label{table:aslsota}
\end{table}

\section{Conclusion} \label{sec:conclusion}

We have demonstrated the advantages
of using visual keyword spotting to automatically
annotate continuous sign language videos
with weakly-aligned subtitles. We have
presented \datasetName{}, a large-scale
dataset of co-articulated signs that,
coupled with a 3D CNN training, allows
high-performance
recognition of signs from a large-vocabulary.
Our model has further shown beneficial
as initialisation for ASL benchmarks.
Finally, we have provided ablations and baselines
for sign recognition
and sign spotting tasks. A potential future
direction is leveraging
our automatic annotations and recognition model
for sign language translation.

\bigskip
\noindent\textbf{Acknowledgements.}
This work was supported by EPSRC grant ExTol. We also thank T.~Stafylakis, A.~Brown, A.~Dutta, L.~Dunbar, A.~Thandavan, C.~Camgoz, O.~Koller, H.~V.~Joze, O.~Kopuklu for their help.

\clearpage
\bibliographystyle{splncs04}
\bibliography{references}

\title{APPENDIX}
\author{}
\institute{}
\maketitle

\renewcommand{\thefigure}{A.\arabic{figure}}
\setcounter{figure}{0} 
\renewcommand{\thetable}{A.\arabic{table}}
\setcounter{table}{0} 

\appendix

This document provides additional results (Section~\ref{app:sec:additionalresults}),
details about the video pose distillation model (Section~\ref{app:sec:posedistillation}),
and about the \datasetName{} dataset (Section~\ref{app:sec:dataset}).

\section{Additional Results} \label{app:sec:additionalresults}
In this section, we present complementary results
to the main paper.
Section~\ref{app:subsec:qualitative} provides a qualitative analysis.
Additional experiments investigate
the search window size for mouthing (Section~\ref{app:subsec:kwswindow}),
the number of frames for sign recognition (Section~\ref{app:subsec:numframes}),
the effect of masking the mouth at test time (Section~\ref{app:subsec:mouthmask}),
ensembling part-specific models (Section~\ref{app:subsec:fusion}),
the transfer to co-articulated datasets (Section~\ref{app:subsec:transfer}),
and the baselines using other cues (Section~\ref{app:subsec:othercues}).

\subsection{Qualitative analysis} \label{app:subsec:qualitative}
\begin{figure}
    \centering
    \includegraphics[width=.99\textwidth,trim={0cm 0.7cm 0cm 1.5cm},clip]{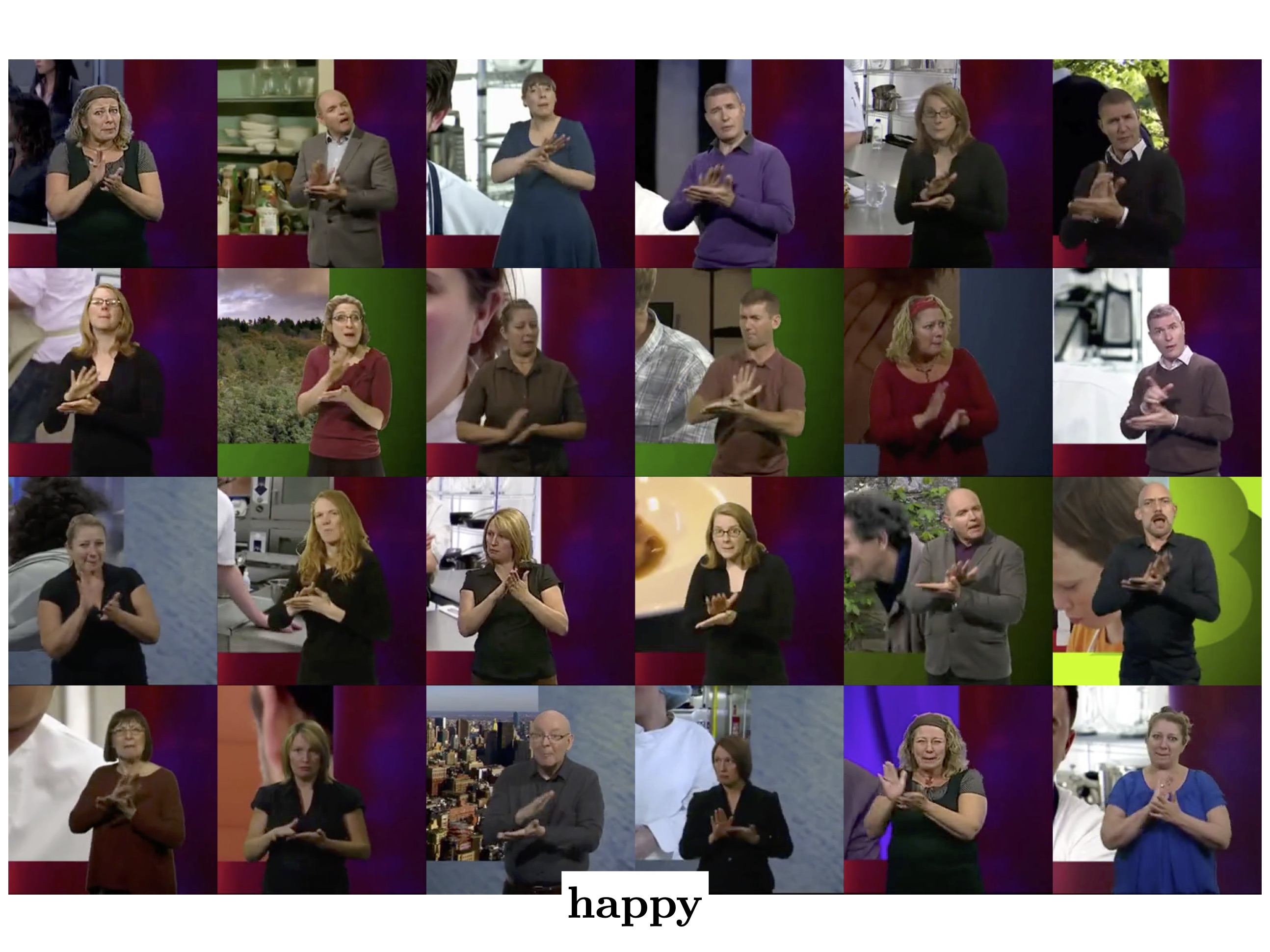}
    \includegraphics[width=.99\textwidth,trim={0cm 0.7cm 0cm 1.5cm},clip]{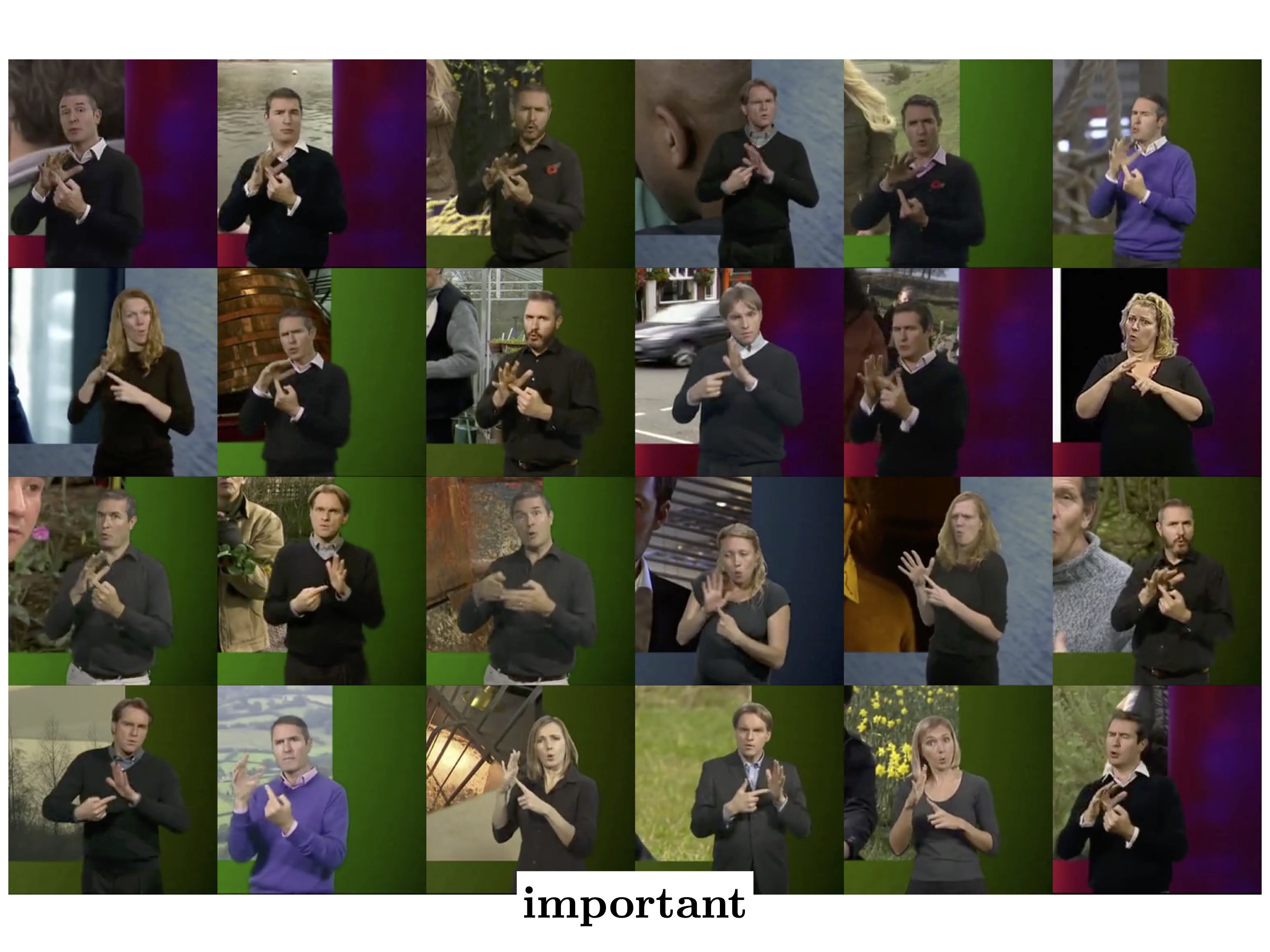}
    \caption{\textbf{Mouthing results:} Qualitative samples for the visual keyword spotting method for the keywords ``happy''
    and ``important''. We visualise the top 24 videos with the
    most confident mouthing scores for each word. We note the visual similarity
    among manual features which suggests that mouthing cues
    can be a good starting point to automatically annotate
    training samples.}
    \label{app:fig:qualmouthing}
\end{figure}

\begin{figure}
    \centering
    \includegraphics[width=.99\textwidth,trim={0cm 0.7cm 0cm 1.5cm},clip]{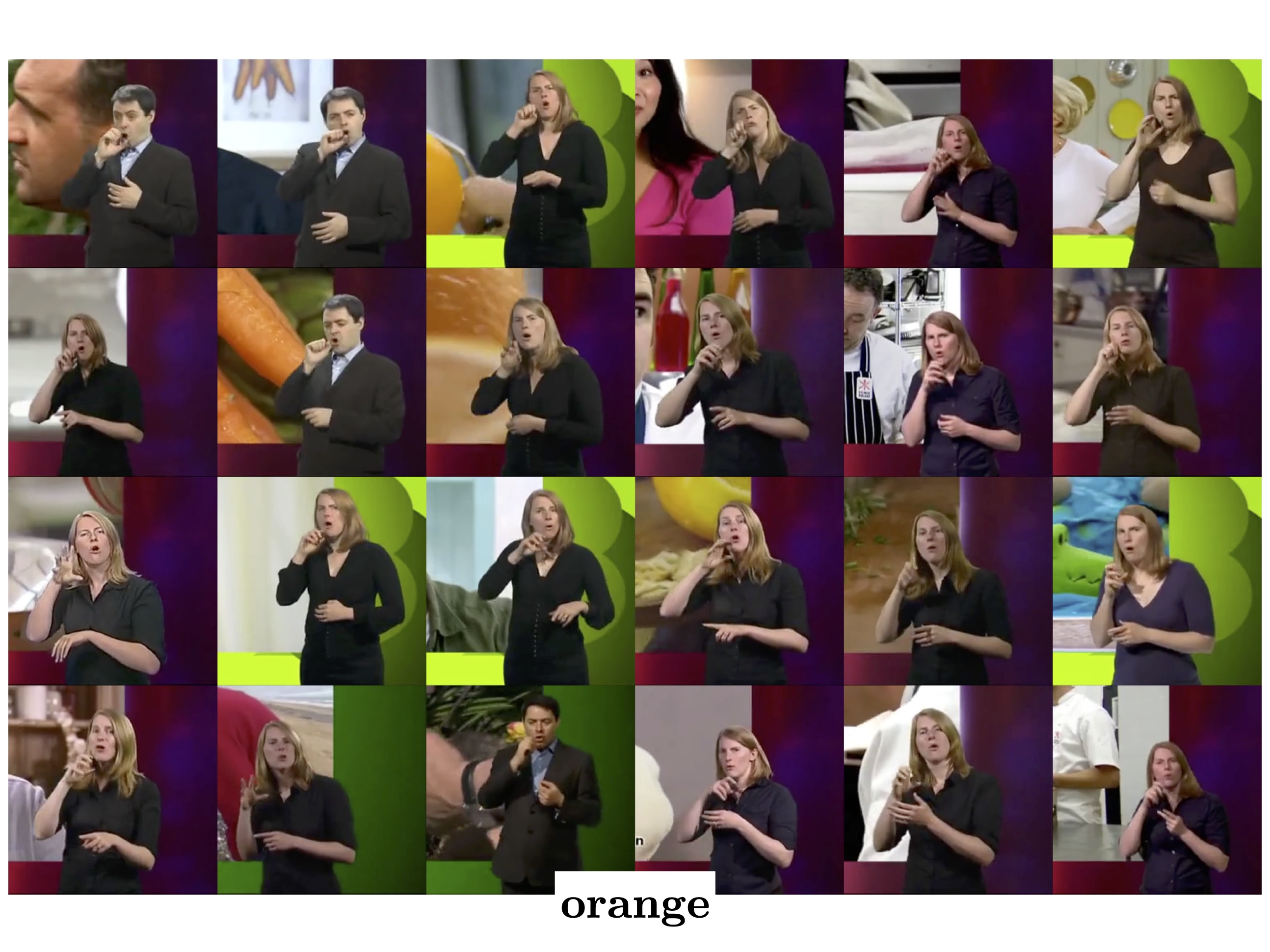}
    \includegraphics[width=.99\textwidth,trim={0cm 0.7cm 0cm 1.5cm},clip]{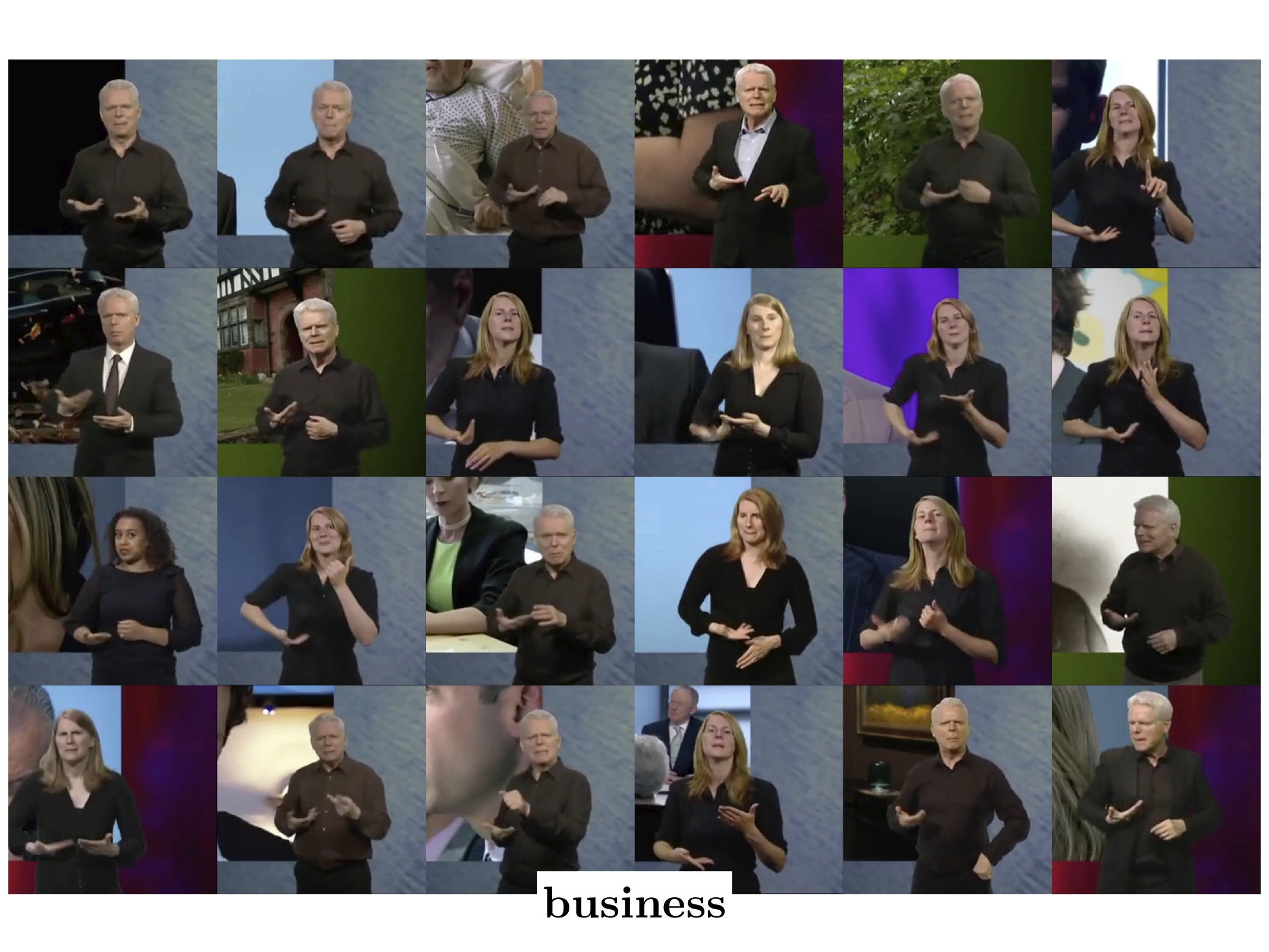}
    \caption{\textbf{Sign recognition results:} Qualitative
    samples for our sign language recognition I3D model on the \datasetName{}
    test set. We visualise the top 24 videos
    with the highest classification scores for the signs
    ``orange'' and ``business'', which appear to be all correctly
    classified. }
    \label{app:fig:qualrecognition}
\end{figure}

We provide a video on our project page\footnote{ \url{https://www.robots.ox.ac.uk/~vgg/research/bsl1k/}}
to illustrate the automatically annotated training samples
in our \datasetName{} dataset, as well
as the results of our sign recognition
model on the manually verified test set.
Figures~\ref{app:fig:qualmouthing}~and~\ref{app:fig:qualrecognition}
present some of these results.
In Figure~\ref{app:fig:qualmouthing}, we provide
\textit{training} samples localised using mouthing cues.
In Figure~\ref{app:fig:qualrecognition}, we provide
\textit{test} samples classified by our I3D model trained
on the automatic annotations.

\subsection{Size of the search window for visual keyword spotting} \label{app:subsec:kwswindow}

We investigate the influence of varying the extent of the temporal window around a given subtitle during the visual keyword spotting phase of dataset collection.  For this experiment, we run the visual keyword spotting model with different search window sizes (centring these windows on the subtitle locations), and train sign recognition models (following the protocol described in the main paper, using Kinetics initialisation) on the resulting annotations.   We find (Table~\ref{app:table:kwswindow}) that 8-second extended search windows yield the strongest performance on the test set (which is fixed across each run)---we therefore use these for all experiments used in the main paper.

\setlength{\tabcolsep}{8pt}
\begin{table}[t]
    \centering
    \caption{The effect of the temporal window where we apply the visual keyword spotting
        model. Networks are trained on \datasetName{}$_{m.8}$ with Kinetics initialisation.
        Decreasing the window size increases the chance of missing the word,
        resulting in less training data and lower performance. Increasing too much makes the
        keyword spotting task difficult, reducing the annotation quality. We found 8 seconds to be
        a good compromise, which we used in all other experiments in this paper.}
    \resizebox{0.8\linewidth}{!}{
        \begin{tabular}{rccccc}
            \toprule
            & & \multicolumn{2}{c}{per-instance} & \multicolumn{2}{c}{per-class} \\
            Keyword search window & \#videos & top-1 & top-5 & top-1 & top-5  \\
            \midrule
            1 sec & 25.0K & 60.10 & 75.42 & 36.62 & 53.83 \\ 
            2 sec & 33.9K & 64.91 & 80.98 & 40.29 & 59.63 \\ 
            4 sec & 37.6K & 68.09 & 82.79 & 45.35 & 63.64 \\ 
            8 sec & 38.9K & \textbf{69.00} & \textbf{83.79} & \textbf{45.86} & \textbf{64.42} \\ 
            16 sec & 39.0K & 65.91 & 81.84 & 39.51 & 59.03 \\ 
            \bottomrule
        \end{tabular}
    }
    \label{app:table:kwswindow}
\end{table}

\subsection{Temporal extent of the automatic annotations} \label{app:subsec:numframes}
Keyword spotting provides a precise localisation in time,
but does not determine the \textit{duration} of the sign. We observe
that the highest mouthing confidence is obtained at the \textit{end}
of mouthing. We therefore take a certain number of frames before
this peak to include in our sign classification training.
In Table~\ref{app:table:duration}, we experiment with this hyper-parameter and see that
20 frames is a good compromise for creating variation in training,
while not including too many irrelevant frames. In all of our
experiments, we used 24 frames, except in
Table~\ref{table:best}
which combines the best parameters from each ablation,
where we used 20 frames.
Note that our I3D
model takes in 16 consecutive frames as input, which is sliced
randomly during training.
\begin{table}[t]
    \centering
    \caption{The effect of the number of frames before the mouthing peak used for training.
        Networks are trained on \datasetName{}$_{m.8}$ with Kinetics initialisation.}
    \resizebox{0.5\linewidth}{!}{
        \begin{tabular}{lcccc}
            \toprule
            & \multicolumn{2}{c}{per-instance} & \multicolumn{2}{c}{per-class} \\
            \#frames & top-1 & top-5 & top-1 & top-5  \\
            \midrule
            16 & 59.53 & 77.08 & 36.16 & 58.43 \\
            20 & \textbf{71.71} & \textbf{85.73} & \textbf{49.64} & \textbf{69.23} \\
            24 & 69.00 & 83.79 & 45.86 & 64.42 \\
            \bottomrule
        \end{tabular}
    }
    \label{app:table:duration}
\end{table}

\subsection{Masking the mouth region at test time} \label{app:subsec:mouthmask}

\begin{figure}[t]
    \centering
    \includegraphics[width=.99\textwidth,trim={0cm 0cm 0cm 0cm},clip]{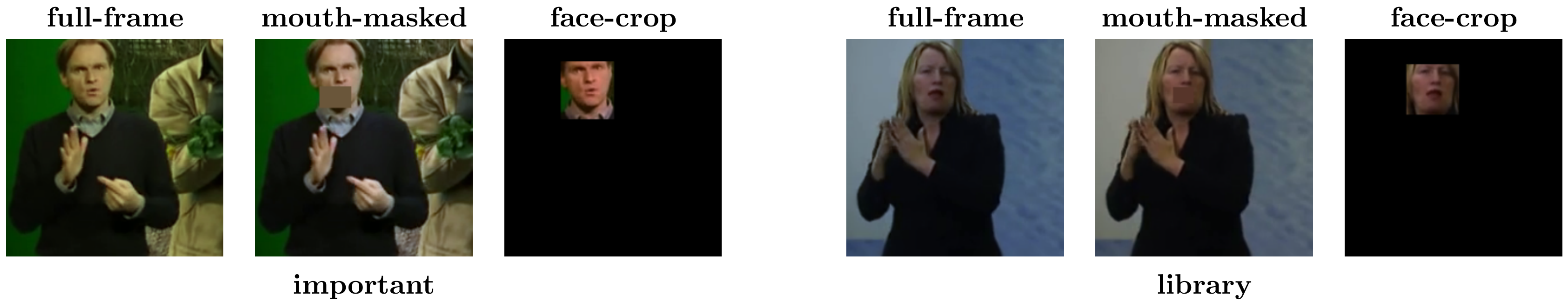}
    \caption{\textbf{Masking the mouth:} Sample visualisations for the inputs
    described in
    Table~\ref{table:pose2sign}
    (for the signs ``important'' and ``library''). We experiment with masking the mouth
    region or cropping only the face region using the detected pose keypoints.
    }
    \label{app:fig:mask}
\end{figure}
In Table~\ref{app:table:pose2signtest},
we experiment with the test modes for the networks trained with
(i) full-frames including the mouth, versus
(ii) masking the mouth region.
If the mouth is masked only
at test time, the performance drops from 65.57\% to 34.74\%
suggesting the model's significant reliance on the mouth cues.
The model can be improved to 46.75\% if it is forced to
focus on other regions by training with masked mouth.

\begin{table}[t]
    \centering
    \caption{We complement
        Table~\ref{table:pose2sign}
        by investigating different test modes for I3D, when trained with
        or without the mouth pixels. The model trained with full-frames
        relies significantly on the mouth, whose performance drops from 65.57\%
        to 34.74\% when the mouth is masked. The models are trained on
        the subset of \datasetName{}$_{m.8}$ where pose estimates are available.
    }
    \resizebox{0.9\linewidth}{!}{
        \begin{tabular}{lcccc|cccc}
            \toprule
            & \multicolumn{4}{c}{Test mouth-masked} & \multicolumn{4}{c}{Test full-frame} \\
            & \multicolumn{2}{c}{per-instance} & \multicolumn{2}{c|}{per-class} & \multicolumn{2}{c}{per-instance} & \multicolumn{2}{c}{per-class} \\
            & top-1 & top-5 & top-1 & top-5 & top-1 & top-5 & top-1 & top-5  \\
            \midrule
            Train mouth-masked & \textbf{46.75} & \textbf{66.34} & \textbf{25.85} & \textbf{48.02} & 46.21 & 65.34 & 25.83 & 46.23 \\
            Train full-frame & 34.74 & 51.42 & 13.62 & 29.80 & \textbf{65.57} & \textbf{81.33} & \textbf{44.90} & \textbf{64.91} \\
            \bottomrule
        \end{tabular}
    }
    \label{app:table:pose2signtest}
\end{table}

\begin{table}[h!]
    \centering
    \caption{Ensembling part-specific models from 
        Table~\ref{table:pose2sign}.
        We observe that combining the I3D model trained only with the face
        and another model without the mouth (last row) achieves
        superior performance than using one model that inputs the
        full-frame. This suggests that disentangling manual and non-manual
        features, which are complementary, for sign recognition is a promising direction.
        The models are trained on
        the subset of \datasetName{}$_{m.8}$ where pose estimates are available.
    }
    \resizebox{0.7\linewidth}{!}{
        \begin{tabular}{llcccc}
            \toprule
                & &\multicolumn{2}{c}{per-instance} & \multicolumn{2}{c}{per-class} \\
                          & & top-1 & top-5 & top-1 & top-5 \\
            \midrule
            face-crop    && 42.23 & 69.70 & 21.66 & 50.51 \\
            mouth-masked && 46.75 & 66.34 & 25.85 & 48.02 \\
            full-frame   && \textbf{65.57} & \textbf{81.33} & \textbf{44.90} & \textbf{64.91} \\
            \bottomrule
            full-frame & \hspace{-0.4cm} \& face-crop & 64.50 & 83.01 & 42.30 & 65.58 \\
            full-frame & \hspace{-0.4cm} \& mouth-masked & 68.09 & 81.33 & \textbf{46.29} & 65.41 \\
            mouth-masked & \hspace{-0.4cm} \& face-crop & \textbf{68.55} & \textbf{83.63} & 45.29 & \textbf{67.47} \\
            \bottomrule
        \end{tabular}
    }
    \label{app:table:poseensembling}
\end{table}

\subsection{Late fusion of part-specific models.} \label{app:subsec:fusion}
We further experiment with ensembling two I3D networks,
each specialising on different parts of the human body,
by averaging the classification scores, i.e., late fusion.
The results are summarised in Table~\ref{app:table:poseensembling}.
We observe significant improvements
when combining a mouth-specific model (face-crop)
with a body-specific model (mouth-masked),
which suggests that forcing the network to
focus on separate, but complementary signing cues (68.55\%)
can be more effective than presenting the full-frames (65.57\%).
This procedure; however, involves additional complexity
of computing the human pose and training two separate models.
It is therefore only used for experimental purposes.

Figure~\ref{app:fig:mask} presents sample
visualisations for the masking procedure.
For mouth-masking, we replace the box covering the mouth region
with the average pixel of the region. For face-cropping,
we set pixels outside of the face region to zero
(we observed divergence of training if the mean value
was used).

\subsection{Transferring BSL-1K pretrained model to other datasets}\label{app:subsec:transfer}
As explained in
Section~\ref{subsec:exp:sota},
we use our model pretrained on BSL-1K as initialisation for transferring
to other datasets.

\noindent\textbf{Additional details on fine-tuning for ASL.}
For MSASL~\cite{Joze19msasl} and WLASL~\cite{Li19wlasl}
isolated datasets on ASL, we have used the pretraining with the
mouth-masking to force the model to entirely pay attention to manual features.
We also observed that some signs are identical between ASL and BSL; therefore,
instead of randomly initialising the last classification layer, we have kept
the weights corresponding to common words between BSL-1K and the ASL dataset.
We observed slight improvements with both of these choices.

\noindent\textbf{Results on co-articulated datasets.}
Here, we report the results
of training sign language recognition on two co-articulated datasets: (i) RWTH-PHOENIX-Weather-2014-T~\cite{Koller15cslr,Camgoz18}
and (ii) BSL-Corpus~\cite{bslcorpus17,schembri2013building}, with and without
pretraining on BSL-1K.

Phoenix dataset is not directly applicable to our model due to the lack of sign-gloss
alignment to train I3D with short clips of individual signs. We therefore
implemented a simple CTC loss \cite{graves2006ctc} to adapt I3D for Phoenix
and obtained 5.6 WER improvement with BSL-1K pretraining over Kinetics pretraining.

BSL-Corpus is a linguistic dataset, and has not been used for computer vision research so far. We defined a train/val/test split (8:1:1 ratio) for a subset of 6k annotations of 966 signs and obtained 24.4\% vs 12.8\% accuracy with/without BSL-1K pretraining. In this case, we have also kept the last-layer classification weights for which the words are in common between BSL-Corpus and BSL-1K signs. We observed this to provide small gains over completely random initialisation of classification weights.

We conclude that our large-scale BSL-1K dataset provides a strong initialisation
for both co-articulated and isolated datasets; for a variety of sign languages: ASL (American),
BSL (British), and DGS (German).

\subsection{Dataset expansion through other cues and additional baselines} \label{app:subsec:othercues}

In addition to the experiments reported in the paper, we further implemented the dataset labelling technique described in~\cite{pfister2013large} which searches subtitles for signs and picks candidate temporal windows that maximise the area under the ROC curve for positively and negatively labelled bags (here, a positive bag refers to temporal windows that occur within an approximately 400 frame interval centred on the target word).   However, we found that without the use of the keyword spotting model for localisation, the annotations collected with this technique were extremely noisy, and the resulting model significantly under-performed all baselines reported in the main paper (that were instead trained on \datasetName).  We also experimented with dataset expansion through training ensembles of exemplar SVMs~\cite{malisiewicz2011ensemble} for each episode on signs that were predicted as confident positives (greater than 0.8) by our strongest pretrained model (and using all temporal windows that did not include the keyword as negatives).  In this case, we found it challenging to calibrate SVM confidences (we explored both the parameters of the original paper, who discuss the difficulties of this process~\cite{malisiewicz2011ensemble} and a range of other parameters) and expanded the dataset by a factor of three, but did not achieve a boost in model performance when training on the expanded data.

\section{Video Pose Distillation} \label{app:sec:posedistillation}

In this section, we give additional details of the video pose distillation model described in
Section~\ref{subsec:method:posedistillation}.
The model architecture uses an I3D backbone~\cite{Carreira2017} and takes as input a sequence of 16 frames at $224\times224$ pixels. We remove the classification head used in the original architecture and replace it with a linear layer that projects the 1024-dimensional embedding to 4160 dimensions---this corresponds to a set of 16 per-frame predictions of the $xy$ coordinates of the 130 human pose keypoints produced by an OpenPose~\cite{cao2018openpose} model (trained on COCO~\cite{lin2014microsoft}).  The coordinates are normalised (with respect to the dimensions of the input image) to lie in the range $[0, 1]$ and an L2 loss is used to penalise inaccurate predictions.  The training data for the pose distillation model comprises one-minute segments from each episode used in the \datasetName{} training set.  The model is trained for 100 epochs using the Lookahead optimizer~\cite{zhang2019lookahead} with minibatches of 32 clips using a learning rate of $0.1$ (reduced by a factor of $10$ after 50 epochs) and a weight decay of $0.001$.

\section{\datasetName{} Dataset Details} \label{app:sec:dataset}
\subsection{Sign verification tool}

\begin{figure}[t]
    \centering
    \includegraphics[width=.8\textwidth,trim={0cm 2cm 20cm 0cm},clip]{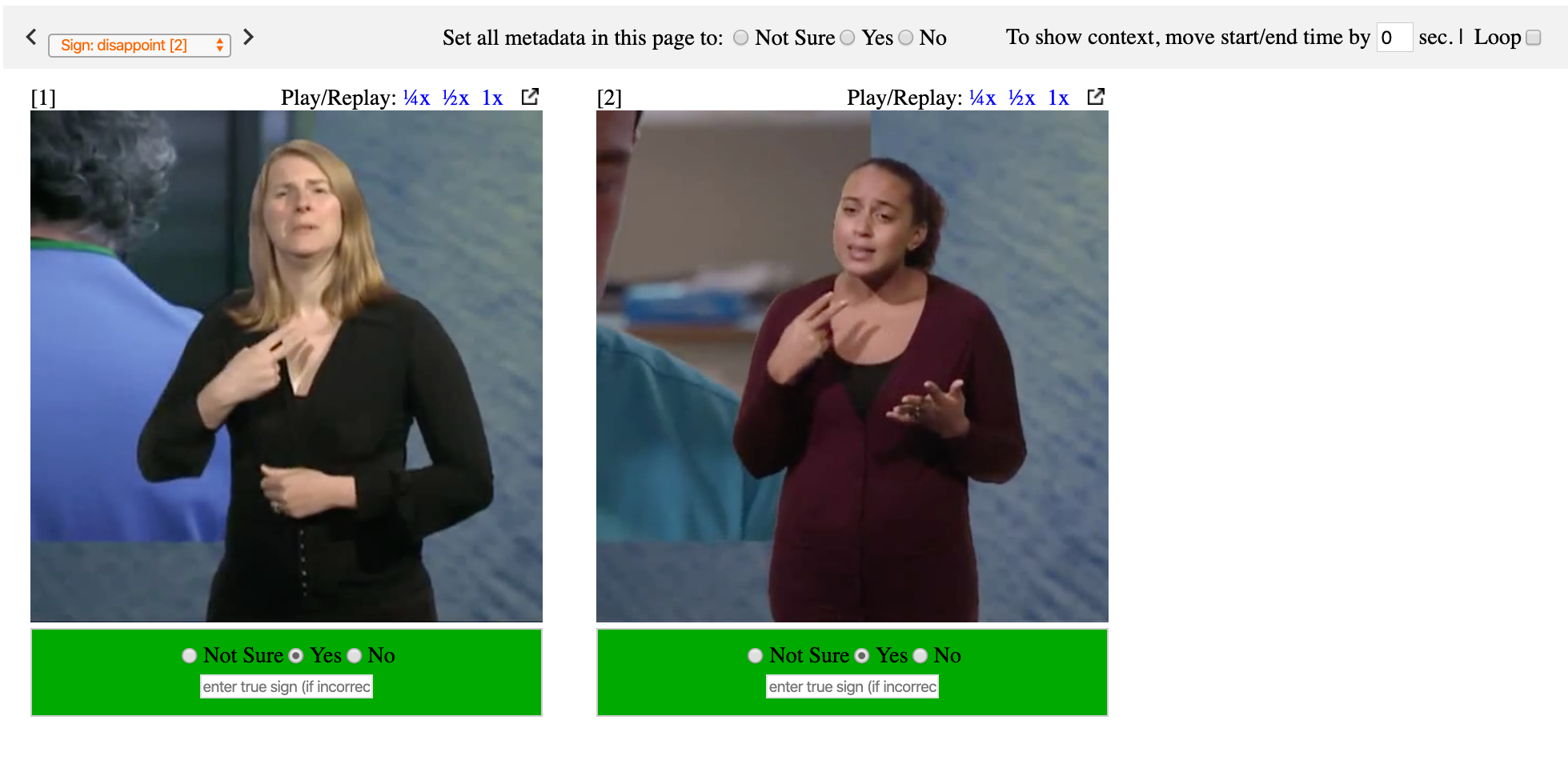}
    \caption{\textbf{Manual annotation:} A screenshot of the Whole-Sign Verification Tool.}
    \label{app:fig:verification-tool}
\end{figure}

In Figure~\ref{app:fig:verification-tool}, we show a screenshot of the verification tool used by annotators to verify or reject signs found by the proposed keyword spotting method in the test set.  Annotators have the ability to view the sign at reduced speed and indicate whether the sign is correct, incorrect, or they are unsure.

\subsection{Dataset source material} \label{app:subsec:episodes}

The BBC broadcast TV shows, together with their respective number of occurrences in the source material used to construct the dataset are: \\

\noindent Countryfile: 266, Natural World: 70, Great British Railway Journeys: 122, Holby City: 261, Junior Masterchef: 24, Junior Bake Off: 22, Hairy Bikers Bakeation: 6, Masterchef The Professionals: 37, Doctor Who Sci Fi: 23, Great British Menu: 110, A To Z Of Cooking: 24, Raymond Blanc Kitchen Secrets: 9, The Apprentice: 88, Country Show Cook Off: 18, A Taste Of Britain: 20, Lorraine Pascale How To Be A: 6, Chefs Put Your Menu Where Your: 13, Simply Nigella: 7, The Restaurant Man: 5, Hairy Bikers Best Of British: 27, Rip Off Britain Food: 20, Our Food Uk 4: 3, Disaster Chefs: 8, Terry And Mason Great Food Trip: 19, Gardeners World: 70, Paul Hollywood Pies Puds: 20, James Martin Food Map Of Britain: 10, Baking Made Easy: 6, Hairy Bikers Northern: 7, Nigel Slater Eating Together: 6, Raymond Blanc How To Cook Well: 6, Great British Food Revival: 17, Great British Bake Off: 28, Two Greedy Italians: 4, Food Fighters: 10, Hairy Bikers Mums Know Best: 9, Hairy Bikers Meals On Wheels: 6, Paul Hollywood Bread 6: 5, Home Comfort At Christmas: 1
\subsection{Dataset vocabulary} \label{app:subsec:vocab}

The 1,064 words which form the vocabulary for \datasetName{} are:\\

\small
\noindent abortion, about, above, absorb, accept, access, act, activity, actually, add, address, advance, advertise, afford, afghanistan, africa, afternoon, again, against, agree, aids, alcohol, all, already, always, amazed, america, angel, angry, animal, answer, anything, anyway, apple, apprentice, approach, april, apron, arch, archery, architect, area, argue, arm, army, around, arrive, arrogant, art, asian, ask, assess, atmosphere, attack, attention, attitude, auction, australia, austria, automatic, autumn, average, award, awful, baby, back, bacon, bad, balance, ball, ballet, balloon, banana, bank, barbecue, base, basketball, bath, battery, beach, beat, because, bedroom, beef, been, before, behind, belfast, belgium, believe, belt, better, big, billion, bingo, bird, birmingham, birthday, biscuit, bite, bitter, black, blackpool, blame, blanket, blind, blonde, blood, blue, boat, body, bomb, bone, bonnet, book, booked, border, boring, born, borrow, boss, both, bottle, boundary, bowl, box, boxing, branch, brave, bread, break, breathe, brick, bridge, brief, brighton, bring, bristol, britain, brother, brown, budget, buffet, build, bulgaria, bull, bury, bus, bush, business, but, butcher, butter, butterfly, buy, by, cabbage, calculator, calendar, call, camel, camera, can, canada, cancel, candle, cap, captain, caravan, cardiff, careful, carpenter, castle, casual, cat, catch, catholic, ceiling, cellar, certificate, chair, chalk, challenge, chance, change, character, charge, chase, cheap, cheat, check, cheeky, cheese, chef, cherry, chicken, child, china, chips, chocolate, choose, christmas, church, city, clean, cleaner, clear, clever, climb, clock, closed, clothes, cloud, club, coffee, coffin, cold, collapse, collect, college, column, combine, come, comedy, comfortable, comment, communicate, communist, community, company, compass, competition, complicated, compound, computer, concentrate, confident, confidential, confirm, continue, control, cook, copy, corner, cornwall, cottage, council, country, course, court, cousin, cover, cracker, crash, crazy, create, cricket, crisps, cross, cruel, culture, cup, cupboard, curriculum, custard, daddy, damage, dance, danger, daughter, deaf, debate, december, decide, decline, deep, degree, deliver, denmark, dentist, depend, deposit, depressed, depth, derby, desire, desk, detail, detective, devil, different, dig, disabled, disagree, disappear, disappoint, discuss, disk, distract, divide, dizzy, doctor, dog, dolphin, donkey, double, downhill, drawer, dream, drink, drip, drive, drop, drunk, dry, dublin, dvd, each, early, east, easter, easy, edinburgh, egypt, eight, elastic, electricity, elephant, eleven, embarrassed, emotion, empty, encourage, end, engaged, england, enjoy, equal, escalator, escape, ethnic, europe, evening, everything, evidence, exact, exchange, excited, excuse, exeter, exhausted, expand, expect, expensive, experience, explain, express, extract, face, factory, fairy, fall, false, family, famous, fantastic, far, farm, fast, fat, father, fault, fax, february, feed, feel, fence, fifteen, fifty, fight, film, final, finance, find, fine, finish, finland, fire, fireman, first, fish, fishing, five, flag, flat, flock, flood, flower, fog, follow, football, for, foreign, forever, forget, fork, formal, forward, four, fourteen, fox, france, free, freeze, fresh, friday, friend, frog, from, front, fruit, frustrated, fry, full, furniture, future, game, garden, general, generous, geography, germany, ginger, girl, give, glasgow, glass, gold, golf, gorilla, gossip, government, grab, grandfather, grandmother, greedy, green, group, grow, guarantee, guess, guilty, gym, hair, half, hall, hamburger, hammer, hamster, handshake, hang, happen, happy, hard, hat, have, headache, hearing, heart, heavy, helicopter, hello, help, hide, history, holiday, holland, home, hope, hopeless, horrible, horse, hospital, hot, hotel, hour, house, how, hundred, hungry, hypocrite, idea, if, ignore, imagine, impact, important, impossible, improve, income, increase, independent, india, influence, inform, information, injection, insert, instant, insurance, international, internet, interrupt, interview, introduce, involve, ireland, iron, italy, jamaica, january, japan, jealous, jelly, jersey, join, joke, jumper, just, kangaroo, karate, keep, kitchen, label, language, laptop, last, late, later, laugh, leaf, leave, leeds, left, lemon, library, lighthouse, lightning, like, line, link, list, little, live, liverpool, london, lonely, lost, love, lovely, machine, madrid, magic, make, man, manage, manchester, many, march, mark, marry, match, maybe, meaning, meat, mechanic, medal, meet, meeting, melon, member, mention, message, metal, mexico, milk, million, mind, minute, mirror, miserable, miss, mistake, mix, monday, money, monkey, month, more, morning, most, mother, motorbike, mountain, mouse, move, mum, music, must, name, nasty, national, naughty, navy, necklace, negative, nervous, never, newcastle, newspaper, next, nice, nightclub, normal, north, norway, not, nothing, notice, november, now, number, nursery, object, october, offer, office, old, on, once, one, onion, only, open, operate, opposite, or, oral, orange, order, other, out, oven, over, overtake, owl, own, pack, page, pager, paint, pakistan, panic, paper, paris, park, partner, party, passport, past, pattern, pay, payment, pence, penguin, people, percent, perfect, perhaps, period, permanent, person, personal, persuade, petrol, photo, piano, picture, pig, pineapple, pink, pipe, place, plain, plan, plaster, plastic, plate, please, plenty, plumber, plus, point, poland, police, politics, pop, popular, porridge, portsmouth, portugal, posh, poster, potato, pound, practise, praise, prefer, pregnant, president, pretend, price, priest, print, prison, problem, professional, professor, profile, profit, project, promise, promote, protestant, proud, provide, pub, pull, pulse, punch, purple, push, pyramid, quality, quarter, question, quick, quiet, quit, rabbit, race, radio, rain, rather, read, reading, ready, really, receipt, receive, recommend, red, reduce, region, regular, relationship, relax, release, relief, remember, remind, remove, rent, repair, replace, research, resign, respect, retire, review, rhubarb, rich, ride, right, river, rocket, roll, roman, room, roots, rough, round, rub, rubbish, rugby, run, russia, sack, sad, safe, same, sand, sandwich, satellite, saturday, sauce, sausage, school, science, scissors, scotland, scratch, search, second, seed, seem, self, sell, send, sense, sensitive, sentence, separate, september, sequence, service, settle, seven, sex, shadow, shakespeare, shame, shark, sharp, sheep, sheet, sheffield, shine, shirt, shop, should, shoulder, shout, show, shower, sick, sight, sign, silver, similar, since, sister, sit, six, size, skeleton, skin, sleep, sleepy, small, smell, smile, snake, soft, some, sometimes, son, soon, sorry, south, spain, specific, speech, spend, spicy, spider, spirit, split, sport, spray, spread, squash, squirrel, staff, stand, star, start, station, still, story, straight, strange, stranger, strawberry, stress, stretch, strict, string, strong, structure, stubborn, stuck, student, study, stupid, success, sugar, summer, sun, sunday, sunset, support, suppose, surprise, swan, swap, sweden, sweep, swim, swing, switzerland, sympathy, take, talk, tap, taste, tax, taxi, teacher, team, technology, television, temperature, temporary, ten, tennis, tent, terminate, terrified, that, theory, therefore, thing, think, thirsty, thousand, three, through, thursday, ticket, tidy, tiger, time, tired, title, toast, together, toilet, tomato, tomorrow, toothbrush, toothpaste, top, torch, total, touch, tough, tournament, towel, town, train, training, tram, trap, travel, tree, trouble, trousers, true, try, tube, tuesday, turkey, twelve, twenty, twice, two, ugly, ultrasound, umbrella, uncle, under, understand, unemployed, union, unit, university, until, up, upset, valley, vegetarian, video, vinegar, visit, vodka, volcano, volunteer, vote, wait, wales, walk, wall, want, war, warm, wash, waste, watch, water, weak, weather, wednesday, week, weekend, well, west, wet, what, wheelchair, when, where, which, whistle, white, who, why, wicked, width, wild, will, win, wind, window, wine, with, without, witness, wolf, woman, wonder, wonderful, wood, wool, work, world, worry, worship, worth, wow, write, wrong, yes, yesterday.

\end{document}